\documentclass[journal,twoside]{IEEEtran}

\usepackage{setspace}
\usepackage{amssymb,amsfonts,latexsym}
\usepackage{graphicx}
\usepackage{subfigure}
\usepackage{psfrag}

\usepackage[ruled,vlined]{algorithm2e}
\usepackage[noend]{algpseudocode}

\usepackage{comment}
\usepackage{cite}
\usepackage{url}
\usepackage{times}
\usepackage{epsfig}
\usepackage{multirow}
\usepackage{longtable}
\usepackage{amsfonts}
\usepackage{amssymb}%
\usepackage{color}
\usepackage{mathrsfs}
\usepackage{bm}
\usepackage{booktabs}
\usepackage{threeparttable}
\usepackage[cmex10]{amsmath}
\usepackage{amsmath}
\usepackage[english]{babel}
\usepackage{makecell}

\ifCLASSINFOpdf
\else
\fi

\title{Spatiotemporal Graph Transformer for Traffic Intelligence in Edge Computing}

\author {Laha Ale,~\IEEEmembership{Senior Member,~IEEE},~Letian Lin,~Na Cao,~Zheng Ma,  and~Peng Yu,~\IEEEmembership{Senior Member,~IEEE}

\thanks{L. Ale is with the School of Computing and Artificial Intelligence, Southwest Jiaotong University, Chengdu, China (e-mail:laha\_ale@swjtu.edu.cn).}

\thanks{L. Ale, L. Lin and  N. Cao are with SWJTU-Leeds Joint School, Southwest Jiaotong University, Chengdu, China (e-mail: \{L.Ale, sc23ll, sc23n2c\}@leeds.ac.uk).}

\thanks{Z. Ma is with the Key Lab of Information Coding and Transmission, Southwest Jiaotong University, Chengdu, China (e-mail: zma@home.swjtu.edu.cn).}

\thanks{P. Yu  is with the State Key Laboratory of Networking and Switching Technology, Beijing University of Posts and Telecommunications,
Beijing, China (e-mail:yupeng@bupt.edu.cn).}

 \thanks{\emph{Corresponding authors: Peng Yu (email: yupeng@bupt.edu.cn) }}

}
\begin{document}

\maketitle

\begin{abstract}

Accurate traffic forecasting is essential for proactive resource management in edge computing, where service demand evolves dynamically across both space and time. In practical cellular edge systems, traffic exhibits strong spatial correlations among neighboring service regions and long-range temporal dependencies driven by user mobility and application behavior. Existing recurrent forecasting approaches can capture short-term dynamics but often struggle to model long-horizon traffic evolution under non-stationary conditions. To address this challenge, we propose a spatiotemporal graph Transformer framework that jointly models spatial interactions and temporal dependencies for traffic forecasting in edge computing. The framework employs graph neural networks to capture spatial correlations among service regions and leverages Transformer-based self-attention to learn long-range temporal patterns from historical traffic observations. By decoupling spatial representation learning from temporal reasoning, the proposed approach provides an effective mechanism for large-scale spatiotemporal traffic modeling. Extensive experiments on a real-world cellular network dataset demonstrate that the proposed graph Transformer consistently outperforms recurrent graph-based baselines, including GCN-RNN, GCN-LSTM, and GCN-GRU models, across multiple forecasting horizons. The resulting forecasts enable more effective proactive resource provisioning and reduce overload risk compared with reactive management strategies. These results highlight the potential of graph-enhanced attention mechanisms for building intelligent and adaptive edge computing systems.

\end{abstract}


\begin{IEEEkeywords}
 Edge Computing, Traffic, Graphic Neural Network
\end{IEEEkeywords}
\IEEEpeerreviewmaketitle

\section{Introduction}

Edge computing has emerged as a key paradigm for supporting latency-sensitive and bandwidth-intensive applications by moving computation, storage, and intelligence closer to end users and devices~\cite{Xu2021,10944303,Ale2024}. By reducing reliance on centralized cloud infrastructures, edge computing enables real-time services such as immersive media, intelligent transportation, industrial automation, and mobile augmented reality~\cite{Patel2014,Deng2021}. The effectiveness of these systems, however, depends not only on the availability of edge resources but also on the ability to provision them proactively across geographically distributed service regions. Since edge servers typically operate under limited and heterogeneous resource constraints~\cite{Deng2021}, anticipating future service demand has become a fundamental requirement for reliable and efficient edge computing.

Recent advances in AI-native networking and 6G systems have highlighted the growing importance of \emph{traffic intelligence}, which aims to understand, predict, and exploit spatiotemporal traffic dynamics for proactive network management. Wireless traffic exhibits strong spatial correlations across neighboring regions and complex temporal dependencies driven by user mobility, application behavior, daily routines, and large-scale events~\cite{Sun2022MobileData}. Consequently, reactive resource management strategies based solely on current observations often lead to resource underutilization, service overload, and degraded quality of service~\cite{Wu2024,Duc2019,Zhang2023}. Accurate traffic intelligence therefore serves as a critical enabler for proactive resource provisioning, service placement, task scheduling, and edge orchestration~\cite{Ale2021}.

To improve traffic prediction accuracy, recent studies have explored both foundation-model-based and distributed learning paradigms. TIDES~\cite{zhangTIDES2025} introduces a traffic intelligence framework that leverages large language models and spatial-temporal alignment mechanisms to capture heterogeneous wireless traffic patterns across urban regions. Meanwhile, Zhang \emph{et al.}\cite{zhangGradient2025} propose a federated traffic prediction framework that combines gradient compression and correlation-driven aggregation to model spatial dependencies while reducing communication overhead. These studies demonstrate that exploiting spatial correlations is essential for scalable traffic intelligence in future wireless and edge systems. However, despite substantial progress in spatial modeling, the challenge of capturing long-range temporal dependencies in highly dynamic and non-stationary traffic sequences remains insufficiently addressed~\cite{Ale8660445}.

Traffic intelligence in edge computing is fundamentally a spatiotemporal learning problem. From the spatial perspective, traffic demand in one service region is influenced by neighboring regions through geographic proximity, user mobility, and shared network conditions\cite{Wang2017Spatiotemporal,Yao2023}. From the temporal perspective, traffic exhibits both short-term fluctuations and long-range patterns, including diurnal cycles, weekly regularities, seasonal trends, and bursty events~\cite{Xu2017,Zhang2018}. Traditional recurrent neural networks (RNNs), including long short-term memory (LSTM) and gated recurrent unit (GRU) architectures~\cite{Elman1990,Hochreiter1997,Chung2014}, have been widely adopted for temporal forecasting. Although effective for short-term prediction, these models rely on sequential hidden-state propagation and often struggle to preserve long-range temporal dependencies under non-stationary traffic conditions~\cite{Wu2019}.

Transformer-based models~\cite{Vaswani2017,Zhou2021Informer} provide a compelling alternative through self-attention mechanisms. Rather than compressing historical observations into recursively updated hidden states, Transformers directly model interactions among different time steps and selectively attend to informative historical patterns. This capability is particularly valuable for traffic intelligence, where future traffic demand may depend simultaneously on recent fluctuations, long-term periodic behaviors, and irregular demand transitions. Recent studies have explored Transformer architectures for spatiotemporal traffic prediction in wireless and urban environments~\cite{Gu2023Spatial-TemporalTransformer,Hu2023Citywide,Gong2025STTF}. Nevertheless, the integration of graph-based spatial modeling and Transformer-based temporal reasoning for traffic intelligence in edge computing remains relatively underexplored, especially when evaluated in terms of proactive resource management effectiveness.

To address this challenge, this paper proposes a spatiotemporal graph Transformer framework for traffic intelligence in edge computing. The framework decouples spatial and temporal learning: a graph neural network aggregates traffic information across edge service regions using geographic and adaptive adjacency relationships, while a Transformer module captures long-range temporal dependencies in the resulting spatial representations. This combination supports multi-horizon traffic forecasting for proactive edge resource management.

We evaluate the resulting graph-convolutional Transformer (GCN-Transformer) model~\cite{Scarselli2009,He2022} on a large-scale real-world cellular network dataset and compare it with GCN-RNN, GCN-LSTM, and GCN-GRU baselines under identical settings. The results show that the proposed model delivers more accurate and stable forecasts, especially at longer horizons and during rapid traffic changes, enabling more effective proactive provisioning and lower overload risk than reactive management strategies.

The main contributions of this paper are summarized as follows:

\begin{itemize}
\item We formulate spatiotemporal traffic intelligence as a fundamental capability for proactive resource management in edge computing and identify long-range temporal reasoning as a key challenge in dynamic traffic environments.

\item We propose a spatiotemporal graph Transformer framework that integrates graph neural networks for spatial dependency modeling with Transformer-based temporal reasoning for multi-horizon traffic forecasting.

\item We conduct extensive evaluations on a real-world cellular network dataset and demonstrate that the proposed framework consistently outperforms recurrent graph-based approaches in forecasting accuracy, prediction stability, and support for proactive resource provisioning.
\end{itemize}

\section{System Model}
\label{sec_system_model}

\begin{figure*}[!t]
 \centering
 \includegraphics[width=7.0in]{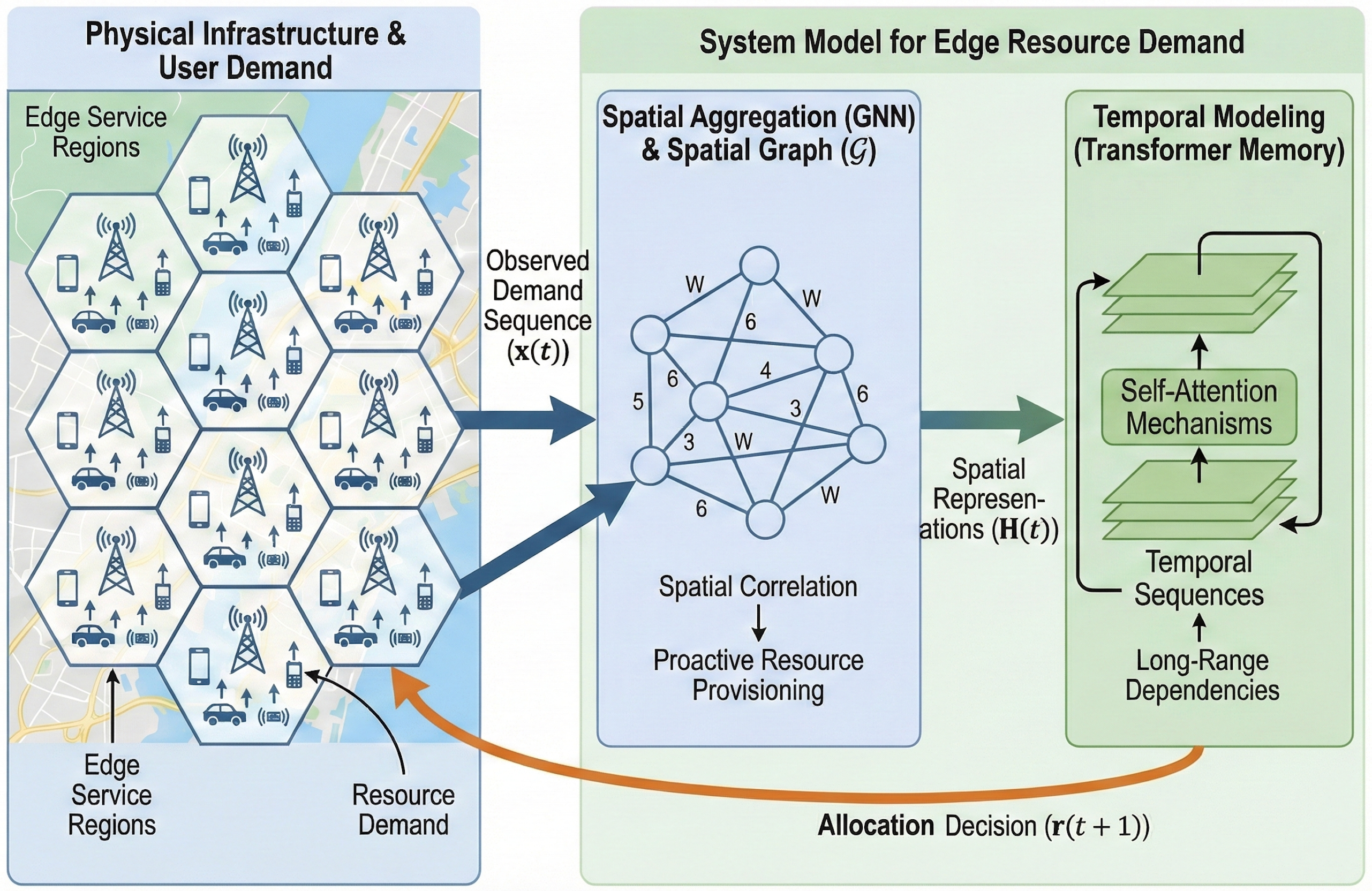}
\caption{Traffic intelligence for proactive resource provisioning in edge computing.}
 \label{fig:system_model}
\end{figure*}

We consider an edge computing system deployed over a large-scale cellular network, as illustrated in Fig.~\ref{fig:system_model}. Edge servers provide computation and storage resources for geographically distributed users. The service area is partitioned into multiple edge service regions, and each region is managed by an edge server or edge controller that provisions resources for nearby devices. The goal of the system is to predict future resource demand in each region and use these predictions to support proactive resource provisioning, thereby reducing overload and service degradation.

\subsection{Spatial Graph Model}

Let $\mathcal{V}=\{1,2,\dots,N\}$ denote the set of edge service regions. Each node $i\in\mathcal{V}$ represents a cluster of geographically proximate base stations or access points. The spatial relationships among regions are represented by a weighted graph $\mathcal{G}=(\mathcal{V},\mathcal{E})$, where $\mathcal{E}\subseteq\mathcal{V}\times\mathcal{V}$ is the set of edges.

An edge $(i,j)\in\mathcal{E}$ indicates that regions $i$ and $j$ are spatially related, for example because they are geographically close, have overlapping service coverage, or are affected by similar user mobility patterns. Each edge is assigned a non-negative weight $w_{ij}$ that measures the strength of this relationship. The resulting weighted adjacency matrix is denoted by $\mathbf{W}\in\mathbb{R}^{N\times N}$ and serves as the graph structure for spatial demand modeling.

\subsection{Traffic Demand Representation}

Time is divided into equal-length intervals indexed by $t=1,2,\dots,T$. Let $x_i(t)$ denote the aggregated resource demand observed in edge service region $i$ during interval $t$. Depending on the application scenario, $x_i(t)$ may represent traffic volume, active service sessions, computation requests, or other indicators of service demand within region $i$.

The network-wide demand state at time $t$ is represented as
\begin{equation}
\mathbf{x}(t) = [x_1(t), x_2(t), \dots, x_N(t)]^\top \in \mathbb{R}^N.
\end{equation}
Given a historical observation window of length $L$, the input demand sequence is
\begin{equation}
\mathcal{X}_t = \{\mathbf{x}(t-L+1),\mathbf{x}(t-L+2),\dots,\mathbf{x}(t)\}.
\end{equation}
The forecasting task is to infer future demand states $\hat{\mathbf{x}}(t+h)$ for one or more prediction horizons $h\geq1$ from $\mathcal{X}_t$ and the spatial graph $\mathcal{G}$.

\subsection{Spatial Feature Aggregation}

Demand in neighboring edge regions is often correlated due to geographic proximity, overlapping coverage, and user mobility. Therefore, before modeling temporal evolution, the system first aggregates spatial information over the graph.

At each time interval $t$, let $\mathbf{X}(t)\in\mathbb{R}^{N\times d_{\mathrm{in}}}$ denote the node-feature matrix, where each row contains the demand-related features of one edge service region and $d_{\mathrm{in}}$ is the input feature dimension. Spatial aggregation propagates information across neighboring nodes according to the weighted graph structure and produces
\begin{equation}
\mathbf{H}(t) = \mathcal{A}\!\left(\mathbf{X}(t),\mathcal{G}\right),
\end{equation}
where $\mathbf{H}(t)\in\mathbb{R}^{N\times d}$ is the spatially contextualized representation and $\mathcal{A}(\cdot)$ denotes a graph-based aggregation operator. The $i$th row of $\mathbf{H}(t)$, denoted by $\mathbf{h}_i(t)$, contains the spatial feature representation of region $i$. The specific parameterization of $\mathcal{A}(\cdot)$ is described in Section~\ref{sec_method}.

\subsection{Temporal Modeling with Transformer Memory}

After spatial aggregation, the temporal evolution of each region is modeled using a Transformer encoder that performs temporal reasoning over historical traffic observations. For region $i$, the sequence of spatial representations
\begin{equation}
\{\mathbf{h}_i(t-L+1), \mathbf{h}_i(t-L+2), \dots, \mathbf{h}_i(t)\}
\end{equation}
is fed into a Transformer encoder. The encoder applies self-attention to learn dependencies among historical time steps, allowing the model to capture both short-term traffic variations and long-range temporal patterns.

The scaled dot-product attention operation is given by
\begin{equation}
\text{Attention}(\mathbf{Q}, \mathbf{K}, \mathbf{V}) = \text{softmax}\!\left(\frac{\mathbf{Q}\mathbf{K}^\top}{\sqrt{d}}\right)\mathbf{V},
\end{equation}
where $\mathbf{Q}$, $\mathbf{K}$, and $\mathbf{V}$ are the query, key, and value projections of the input sequence, respectively. By attending over the historical window, the Transformer provides a context-aware temporal representation for multi-horizon traffic forecasting.

\subsection{Traffic Forecasting and Proactive Resource Provisioning}

The output of the Transformer module is mapped to future traffic forecasts for all edge service regions. These forecasts constitute the traffic intelligence available to the edge controller and are subsequently used to support proactive resource provisioning decisions. By anticipating future traffic demand before it materializes, the controller can allocate computation, storage, and networking resources in advance. Compared with reactive approaches that respond only after demand has been observed, proactive provisioning improves resource utilization, enhances service reliability, and mitigates the impact of sudden traffic surges and spatially correlated overload events.

\section{Problem Formulation}
\label{sec_problem}

Based on the system model introduced in Section~\ref{sec_system_model}, we formulate spatiotemporal traffic forecasting as a supervised learning problem over a graph-structured edge computing system. Given historical traffic observations and the spatial relationships among service regions, the objective is to predict future traffic demand over multiple forecasting horizons. The resulting forecasts provide predictive information for proactive resource provisioning and traffic-aware edge management.

\subsection{Multi-Horizon Spatiotemporal Traffic Forecasting}

Let $\mathcal{G}=(\mathcal{V},\mathcal{E})$ denote the spatial graph of edge service regions. At time interval $t$, the network-wide traffic state is represented by
\begin{equation}
\mathbf{x}_t =
[x_{1,t},x_{2,t},\ldots,x_{N,t}]^\top \in \mathbb{R}^{N},
\end{equation}
where $x_{i,t}$ denotes the traffic demand observed in region $i$, and $N$ is the number of service regions.

Given a historical observation window of length $L$, the input sequence is
\begin{equation}
\mathbf{X}_{t-L+1:t}
=
[\mathbf{x}_{t-L+1},\mathbf{x}_{t-L+2},\ldots,\mathbf{x}_{t}]
\in \mathbb{R}^{N\times L}.
\end{equation}

The goal is to learn a forecasting function $f_{\theta}(\cdot)$ that maps the historical traffic sequence and spatial graph to a sequence of future traffic states:
\begin{equation}
\hat{\mathbf{X}}_{t+1:t+H}
=
f_{\theta}
\left(
\mathbf{X}_{t-L+1:t},
\mathcal{G}
\right),
\end{equation}
where
\begin{equation}
\hat{\mathbf{X}}_{t+1:t+H}
=
[\hat{\mathbf{x}}_{t+1},\hat{\mathbf{x}}_{t+2},\ldots,\hat{\mathbf{x}}_{t+H}]
\in \mathbb{R}^{N\times H}.
\end{equation}

For each prediction horizon $h\in\{1,2,\ldots,H\}$, the model outputs
\begin{equation}
\hat{\mathbf{x}}_{t+h}
=
[\hat{x}_{1,t+h},\hat{x}_{2,t+h},\ldots,\hat{x}_{N,t+h}]^\top .
\end{equation}

Given a training index set $\mathcal{T}$, the model parameters are learned by minimizing the average forecasting error across all training samples, service regions, and prediction horizons. Using the mean absolute error (MAE), the objective is
\begin{equation}
\theta^{\ast}
=
\arg\min_{\theta}
\frac{1}{|\mathcal{T}|HN}
\sum_{t\in\mathcal{T}}
\sum_{h=1}^{H}
\left\|
\hat{\mathbf{x}}_{t+h}
-
\mathbf{x}_{t+h}
\right\|_{1}.
\label{eq:forecast_objective}
\end{equation}

This formulation requires the model to capture spatial correlations among service regions and temporal dependencies across both short-term and long-term traffic patterns.

\subsection{Traffic-Aware Proactive Resource Provisioning}

The predicted traffic states can be used by an edge controller to support proactive resource provisioning. Let
\begin{equation}
\mathbf{r}_{t+h}
=
[r_{1,t+h},r_{2,t+h},\ldots,r_{N,t+h}]^\top
\end{equation}
denote the resource allocation vector at horizon $h$, where $r_{i,t+h}$ represents the computational, storage, or networking resources assigned to region $i$.

A generic traffic-aware provisioning policy can be written as
\begin{equation}
\mathbf{r}_{t+h}
=
g(\hat{\mathbf{x}}_{t+h}),
\end{equation}
where $g(\cdot)$ maps forecasted traffic demand to resource allocation decisions subject to system constraints. For example, regions predicted to experience increased traffic may receive additional resources in advance, while excess resources can be reassigned from regions with lower predicted demand.

The design of the provisioning policy $g(\cdot)$ is beyond the scope of this paper. Instead, this work focuses on improving the accuracy and stability of the spatiotemporal traffic forecasts, which serve as the input to proactive resource management mechanisms.

\section{Proposed Graph-Based Transformer Memory Model}
\label{sec_method}

This section presents the proposed graph-based Transformer memory model for traffic intelligence in edge computing. Following the formulation in Section~\ref{sec_problem}, the model predicts future traffic demand over multiple horizons by combining graph-based spatial aggregation with attention-based temporal memory. The design separates spatial perception from temporal reasoning: the graph module extracts spatially contextualized traffic representations, while the Transformer module reasons over their historical evolution.

\subsection{Model Overview}

The proposed framework consists of three main stages. First, edge service regions are represented as nodes in a weighted spatial graph constructed from geographic proximity. Second, a graph neural network (GNN) aggregates information from neighboring regions to capture spatial traffic correlations. Third, a Transformer-based temporal memory module processes the sequence of spatial representations and produces multi-horizon traffic forecasts for all regions.

At each time interval, node-level traffic features are processed by the GNN through message passing over the spatial graph. The resulting spatial embeddings are arranged as temporal sequences and passed to the Transformer memory, which attends to relevant historical time steps when predicting future traffic demand. The final forecasts can then be used by an edge controller for proactive resource provisioning.

\begin{figure*}[!t]
 \centering
 \includegraphics[width=7.0in]{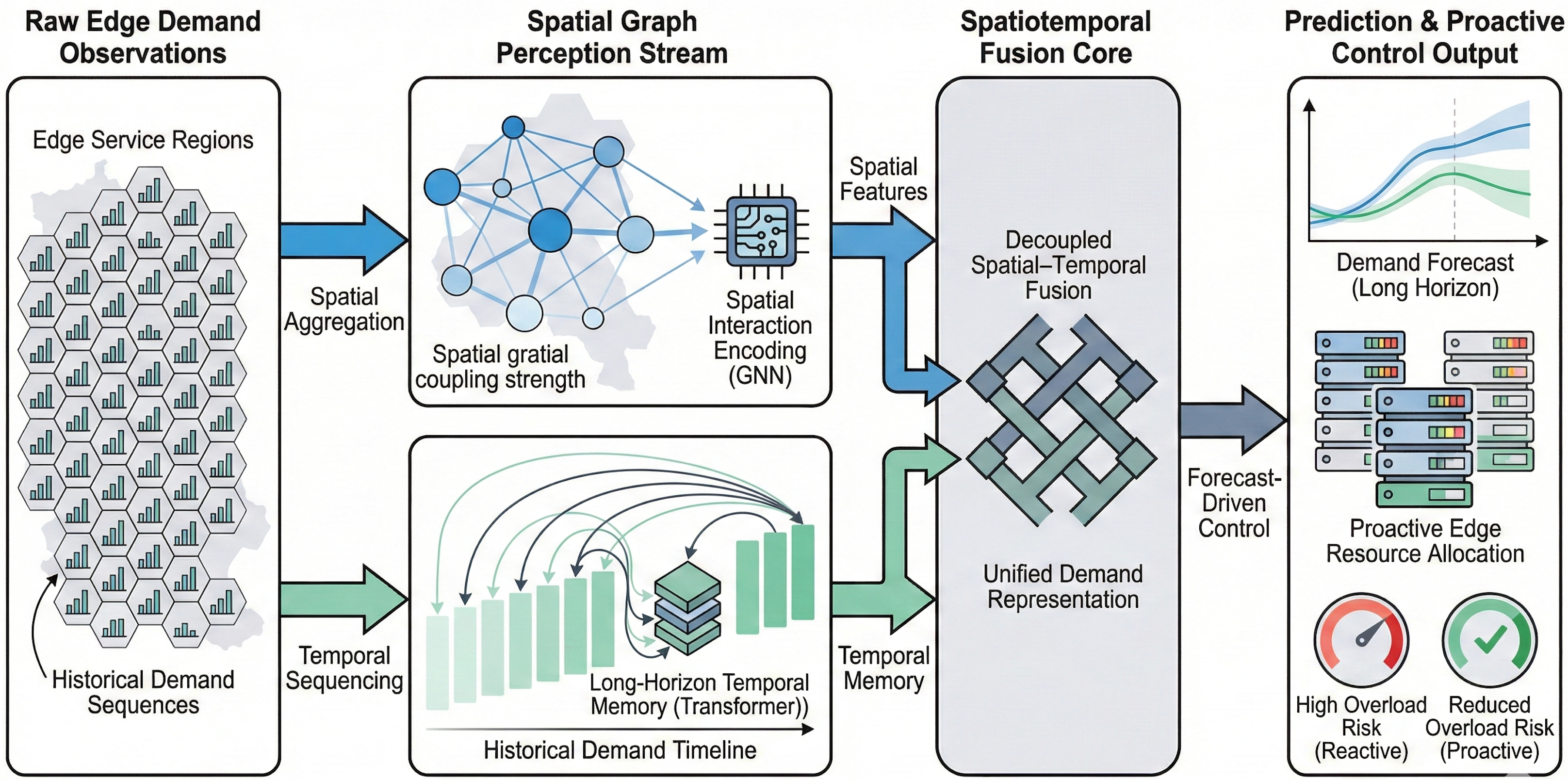}
 \caption{Graph-based Transformer memory framework for traffic intelligence in edge computing.}
 \label{fig:method}
\end{figure*}

Formally, for a historical window $\{\mathbf{X}(t-L+1),\dots,\mathbf{X}(t)\}$ and spatial graph $\mathcal{G}$, the model can be written as
\begin{equation}
\begin{aligned}
\mathbf{H}(\tau) &= \mathcal{F}_{\mathrm{GNN}}\!\left(\mathbf{X}(\tau),\mathcal{G}\right), \quad \tau=t-L+1,\dots,t, \\
\{\hat{\mathbf{x}}(t+h)\}_{h=1}^{H} &=
\mathcal{F}_{\mathrm{out}}\!\left(
\mathcal{F}_{\mathrm{TM}}\!\left(\mathbf{H}(t-L+1),\dots,\mathbf{H}(t)\right)
\right),
\end{aligned}
\end{equation}
where $\mathcal{F}_{\mathrm{GNN}}(\cdot)$ denotes graph-based spatial aggregation, $\mathcal{F}_{\mathrm{TM}}(\cdot)$ denotes Transformer-based temporal memory, and $\mathcal{F}_{\mathrm{out}}(\cdot)$ maps the learned spatiotemporal representation to $H$ future traffic-demand vectors.

\subsection{Input Representation and Spatial Graph Construction}

At each time step $t$, the observed traffic demand across all edge service regions is represented by $\mathbf{x}(t)\in\mathbb{R}^{N}$. This traffic-demand vector can be augmented with contextual features, such as hour-of-day and day-of-week indicators, to form the node-level input matrix
\begin{equation}
\mathbf{X}(t) \in \mathbb{R}^{N \times d_{\mathrm{in}}},
\end{equation}
where each row contains the feature vector of one edge service region and $d_{\mathrm{in}}$ is the input feature dimension.

Spatial interactions are modeled using a weighted graph $\mathcal{G}=(\mathcal{V},\mathcal{E})$. Each node $i\in\mathcal{V}$ corresponds to an edge service region, which is obtained by clustering geographically proximate base stations. Let
\begin{equation}
\mathbf{c}_i=(\phi_i,\lambda_i)
\end{equation}
denote the centroid of region $i$, where $\phi_i$ and $\lambda_i$ are the latitude and longitude of the centroid, respectively. The spatial distance between regions $i$ and $j$ is computed using the Euclidean distance between their centroids:
\begin{equation}
d_{ij}
=
\left\|\mathbf{c}_i-\mathbf{c}_j\right\|_2
=
\sqrt{(\phi_i-\phi_j)^2+(\lambda_i-\lambda_j)^2}.
\end{equation}

A fully connected weighted graph is then constructed over the region centroids. For each pair of regions $(i,j)$, the spatial weight is computed using a Gaussian kernel:
\begin{equation}
w_{ij}
=
\exp\left(
-\frac{d_{ij}^{2}}{\sigma^{2}}
\right),
\end{equation}
where $\sigma>0$ is the kernel bandwidth parameter controlling the spatial decay rate. A smaller $\sigma$ assigns relatively large weights only to nearby regions, while a larger $\sigma$ allows more distant regions to retain stronger spatial correlations. Since the weight decreases as the distance increases, geographically close regions have stronger graph connections.

The resulting weights are assembled into the weighted adjacency matrix $\mathbf{W}\in\mathbb{R}^{N\times N}$, where $W_{ij}=w_{ij}$. The adjacency matrix is then row-normalized as
\begin{equation}
A_{ij}
=
\frac{W_{ij}}{\sum_{k=1}^{N} W_{ik}+\epsilon},
\end{equation}
where $\epsilon$ is a small constant for numerical stability. The normalized adjacency matrix $\mathbf{A}$ is used as the fixed spatial propagation matrix in the graph convolution module. In addition, a learnable adaptive adjacency component is introduced in the model to complement the fixed geographic graph and capture data-driven spatial dependencies that may not be fully explained by physical distance alone.

\begin{algorithm}[!t]
\caption{Spatiotemporal Graph Transformer for Traffic Intelligence in Edge Computing}
\label{alg:gtm_edge}
\DontPrintSemicolon
\KwIn{
Event log $\mathcal{D}$; time slot $\Delta$; history length $L$; forecast horizon $H$;\;
number of regions $N$; Gaussian kernel bandwidth $\sigma$;\;
learning rate $\eta$; batch size $B$; epochs $E$.
}
\KwOut{
Trained parameters $\Theta^\star$; traffic forecasts $\{\hat{\mathbf{x}}(t{+}h)\}_{h=1}^{H}$.
}

\BlankLine
\textbf{Offline preprocessing:}\;
Cluster base-station locations into $N$ edge regions and assign events to regions.\;
Discretize time into fixed-length slots and aggregate node-level traffic demand $\mathbf{x}(t)$ for each region.\;
Construct a fully connected spatial graph over region centroids.\;
Compute pairwise centroid distances using Euclidean distance:
$d_{ij}=\|\mathbf{c}_i-\mathbf{c}_j\|_2$.\;
Compute pairwise edge weights using the Gaussian kernel:
$w_{ij}=\exp(-d_{ij}^{2}/\sigma^{2})$.\;
Form the weighted adjacency matrix $\mathbf{W}$ and obtain the row-normalized adjacency matrix $\mathbf{A}$.\;
Construct sliding-window samples
$\big(\mathbf{X}(t{-}L{+}1{:}t), \{\mathbf{x}(t{+}h)\}_{h=1}^{H}\big)$.\;

\BlankLine
\textbf{Model:}\;
\textit{Spatial aggregation:}
compute $\mathbf{H}(\tau) \leftarrow \sigma_g(\mathbf{A}\mathbf{X}(\tau)\boldsymbol{\Theta}_s)$ for $\tau=t{-}L{+}1,\dots,t$.\;
\textit{Temporal memory:}
apply a Transformer encoder to $\{\mathbf{H}(t{-}L{+}1),\dots,\mathbf{H}(t)\}$.\;
\textit{Output:}
predict $\{\hat{\mathbf{x}}(t{+}h)\}_{h=1}^{H}$.\;

\BlankLine
\textbf{Training:}\;
Initialize model parameters $\Theta$.\;
\For{$e=1$ \KwTo $E$}{
  \ForEach{mini-batch $\mathcal{B}$ of size $B$}{
    Compute spatial embeddings and temporal representations.\;
    Minimize the multi-horizon MAE loss:
    \[
    \mathcal{L}
    =
    \frac{1}{|\mathcal{B}|H}
    \sum_{t\in\mathcal{B}}
    \sum_{h=1}^{H}
    \left\|
    \hat{\mathbf{x}}(t{+}h)-\mathbf{x}(t{+}h)
    \right\|_1 .
    \]
    Update $\Theta \leftarrow \Theta - \eta\nabla_\Theta \mathcal{L}$.\;
  }
}
Set $\Theta^\star \leftarrow \Theta$.\;

\BlankLine
\textbf{Inference:}\;
Given recent observations $\mathbf{X}(t{-}L{+}1{:}t)$, predict future traffic demand
$\{\hat{\mathbf{x}}(t{+}h)\}_{h=1}^{H}$ for all edge service regions.\;
The predicted traffic demand can be used by an edge controller as input for proactive resource provisioning decisions.\;

\end{algorithm}

\subsection{Spatial Modeling via Graph Neural Networks}

To model spatial dependencies among edge service regions, we apply graph convolution over the normalized spatial adjacency matrix. The fixed adjacency matrix $\mathbf{A}^{(0)}$ is obtained from the row-normalized Gaussian kernel weights. In the neural model, this fixed geographic adjacency is further combined with a learnable adaptive adjacency matrix $\mathbf{A}^{(\mathrm{adp})}$:
\begin{equation}
\bar{\mathbf{A}}
=
\mathrm{softmax}
\left(
\mathbf{A}^{(0)}+\mathbf{A}^{(\mathrm{adp})}
\right),
\end{equation}
where the softmax operation is applied row-wise. At each time step $t$, spatial aggregation is performed as
\begin{equation}
\mathbf{H}(t)
=
\rho\left(
\bar{\mathbf{A}}\mathbf{X}(t)\boldsymbol{\Theta}_s
\right),
\end{equation}
where $\boldsymbol{\Theta}_s$ is a learnable weight matrix, $\rho(\cdot)$ is a nonlinear activation function, and $\mathbf{H}(t)\in\mathbb{R}^{N\times d_s}$ is the spatial embedding matrix.

\subsection{Temporal Sequence Construction}

The spatial embeddings obtained from the GNN are arranged into temporal sequences. For each region $i$, the length-$L$ sequence is constructed as
\begin{equation}
\mathbf{S}_i(t) = \left[\mathbf{h}_i(t-L+1);\dots;\mathbf{h}_i(t)\right] \in \mathbb{R}^{L\times d_s},
\end{equation}
where $\mathbf{h}_i(t)$ denotes the $i$th row of $\mathbf{H}(t)$. Thus, $\mathbf{S}_i(t)$ captures the temporal evolution of region $i$ after spatial information from neighboring regions has been incorporated.

\subsection{Transformer-Based Temporal Memory}

To model long-range temporal dependencies, we use a Transformer encoder as the temporal memory module. Unlike recurrent architectures that pass information through a hidden state, the Transformer directly attends to all time steps within the historical window, enabling more flexible temporal reasoning.

Each temporal sequence $\mathbf{S}_i(t)$ is projected into a latent space and augmented with positional encodings to preserve temporal order. Let $\mathbf{U}_i(t)\in\mathbb{R}^{L\times d}$ denote the resulting Transformer input. Multi-head self-attention is computed as
\begin{equation}
\mathrm{MultiHead}(\mathbf{Q},\mathbf{K},\mathbf{V}) =
\mathrm{Concat}(\mathrm{head}_1,\dots,\mathrm{head}_M)\mathbf{W}^{O},
\end{equation}
where $M$ is the number of attention heads and each head is given by
\begin{equation}
\mathrm{head}_m =
\mathrm{softmax}\!\left(\frac{\mathbf{Q}_m\mathbf{K}_m^{\top}}{\sqrt{d_k}}\right)\mathbf{V}_m.
\end{equation}
Here, $\mathbf{Q}_m$, $\mathbf{K}_m$, and $\mathbf{V}_m$ are the query, key, and value projections for head $m$, and $d_k$ is the key dimension. This attention mechanism enables the model to emphasize historical patterns that are most relevant to future traffic demand, including periodic trends and abrupt workload changes.

\subsection{Traffic Forecasting Layer}

Let $\mathbf{Z}_i(t)$ denote the Transformer output sequence for region $i$. A temporal pooling operator $\mathcal{P}(\cdot)$ summarizes this sequence into a compact representation
\begin{equation}
\mathbf{z}_i(t)=\mathcal{P}\!\left(\mathbf{Z}_i(t)\right).
\end{equation}
A fully connected forecasting layer then maps this representation to the multi-horizon prediction vector
\begin{equation}
\hat{\mathbf{y}}_i(t)=\mathbf{W}_o\mathbf{z}_i(t)+\mathbf{b}_o,
\end{equation}
where $\hat{\mathbf{y}}_i(t)=[\hat{x}_i(t+1),\dots,\hat{x}_i(t+H)]^{\top}\in\mathbb{R}^{H}$, and $\mathbf{W}_o$ and $\mathbf{b}_o$ are learnable output parameters. Stacking $\hat{\mathbf{y}}_i(t)$ over all regions gives the full set of forecasts $\{\hat{\mathbf{x}}(t+h)\}_{h=1}^{H}$.

\subsection{Model Training}

The proposed model is trained end-to-end by minimizing the multi-horizon traffic prediction loss defined in Section~\ref{sec_problem}. During training, historical traffic sequences are sampled using a sliding-window strategy. For a mini-batch $\mathcal{B}$, the loss can be written as
\begin{equation}
\mathcal{L}(\Theta)
=
\frac{1}{|\mathcal{B}|H}
\sum_{t\in\mathcal{B}}
\sum_{h=1}^{H}
\left\|
\hat{\mathbf{x}}(t+h)-\mathbf{x}(t+h)
\right\|_1 .
\end{equation}
where $\Theta$ denotes all learnable parameters. Gradients are propagated through both the spatial GNN and the Transformer memory module, enabling joint optimization of spatial perception and temporal reasoning.

\subsection{Computational Considerations}

The proposed architecture is scalable for traffic intelligence in the evaluated cluster-level edge network. Graph convolution is applied to a fully connected weighted adjacency matrix over the $N$ edge regions, which preserves pairwise geographic correlations among all region centroids. Since the experimental graph contains only 25 clusters, the quadratic number of pairwise edges remains computationally manageable. The Transformer memory is applied to fixed-length historical windows and can be parallelized across regions. This design makes the framework suitable for practical edge controllers when the number of regions and the history length are moderate.

\section{Performance Evaluation}
\label{sec_sim_res}

This section evaluates the proposed graph-convolutional Transformer model for traffic intelligence in edge computing. The evaluation focuses on traffic forecasting accuracy across multiple prediction horizons, training and testing behavior, explained variance, node-level prediction trajectories, and the spatial structure of prediction errors. All compared models use the same data split, input window, prediction horizon, and spatial graph representation so that differences can be attributed mainly to the temporal modeling component.

\subsection{Experimental Configuration}
\label{sec_setup}

\subsubsection{Implementation Environment}

The experiments are implemented in Python using PyTorch for model training. Data preprocessing and numerical analysis are performed with NumPy, pandas, SciPy, and Scikit-learn. Visualization and geospatial processing are supported by Matplotlib, Seaborn, Shapely, PyProj, and Cartopy. This software stack is used consistently for all compared models.

\subsubsection{Dataset}

The experiments use the China Telecom Shanghai dataset\footnote{Available at: \url{http://sguangwang.com/TelecomDataset.html}}, a public cellular network dataset collected in Shanghai over a six-month period. The raw dataset contains approximately $6.9\times10^6$ connection records associated with 3,233 base stations. Each record includes user connection information, base-station coordinates, and connection timestamps. The dataset is suitable for traffic demand forecasting in edge computing because it contains both spatial heterogeneity in base-station deployment and temporal variation in user activity.

\subsubsection{Data Preprocessing}

The raw records are transformed into a structured cluster-level time series through four main steps. First, records outside the Shanghai metropolitan region are removed using the longitude range $[120.85^\circ\mathrm{E},122.20^\circ\mathrm{E}]$ and latitude range $[30.70^\circ\mathrm{N},31.90^\circ\mathrm{N}]$. Empty columns are removed, timestamps are converted to datetime format, invalid sessions with negative duration are filtered out, and duplicated sessions are discarded.

\begin{figure}
 \centering
 \includegraphics[width=3.5in]{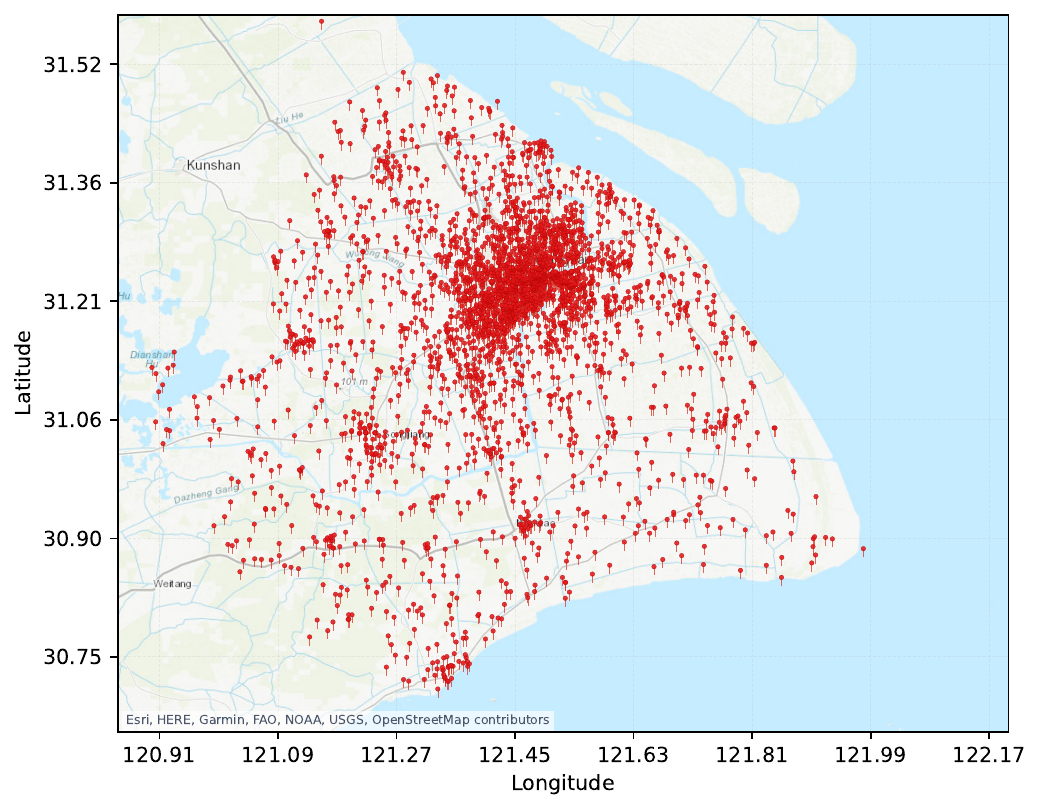}
 \caption{Spatial distribution of the 3,009 valid base-station sites retained after preprocessing in the Shanghai metropolitan area.}
 \label{fig:shanghai_base_stations}
\end{figure}

Second, unique base-station locations are extracted by de-duplicating geographic coordinates, resulting in 3,009 valid spatial sites. Figure~\ref{fig:shanghai_base_stations} shows the spatial distribution of these sites. Base stations are more concentrated in the urban core and become sparser toward peripheral areas, which is consistent with the heterogeneous traffic demand expected in a metropolitan cellular network.

Third, $K$-means clustering is applied to the valid base-station coordinates to form $K_c=25$ edge service regions. Each base station is assigned a \texttt{site\_id} and a \texttt{cluster\_id}, and each cluster is represented by its centroid and the number of base stations it contains. These clusters serve as the graph nodes used in the forecasting model.

\begin{figure*}
 \centering
 \includegraphics[width=7.5in]{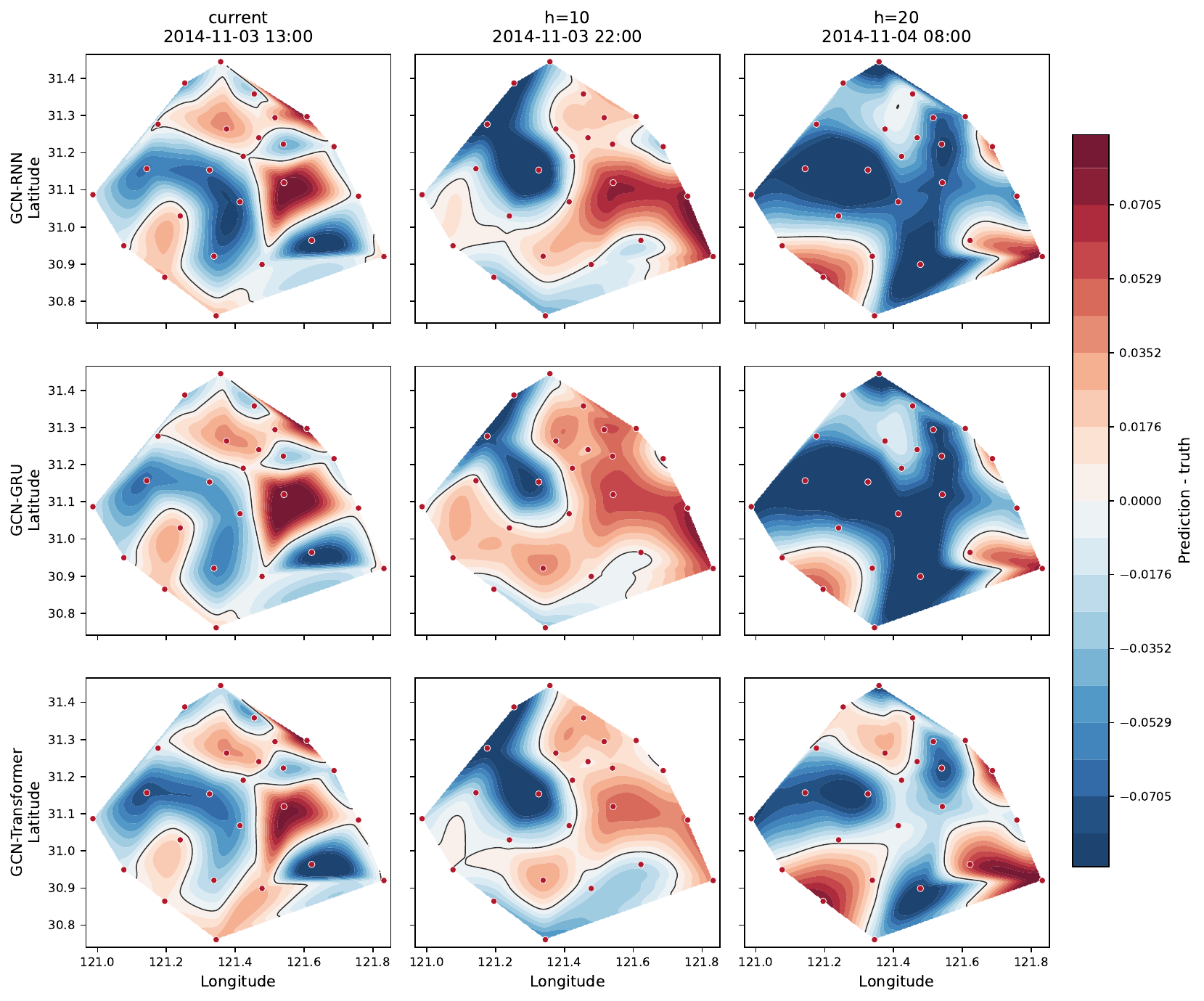}
 \caption{Spatial residual maps for GCN-RNN, GCN-GRU, and GCN-Transformer. Rows correspond to models and columns correspond to the current timestamp, $h{=}10$, and $h{=}20$. Residuals are computed as prediction minus ground truth; red indicates over-prediction, blue indicates under-prediction, and dots mark cluster centroids.}
 \label{fig:residual_heatmap_current}
\end{figure*}

\begin{figure*}
 \centering
 \includegraphics[width=7.5in]{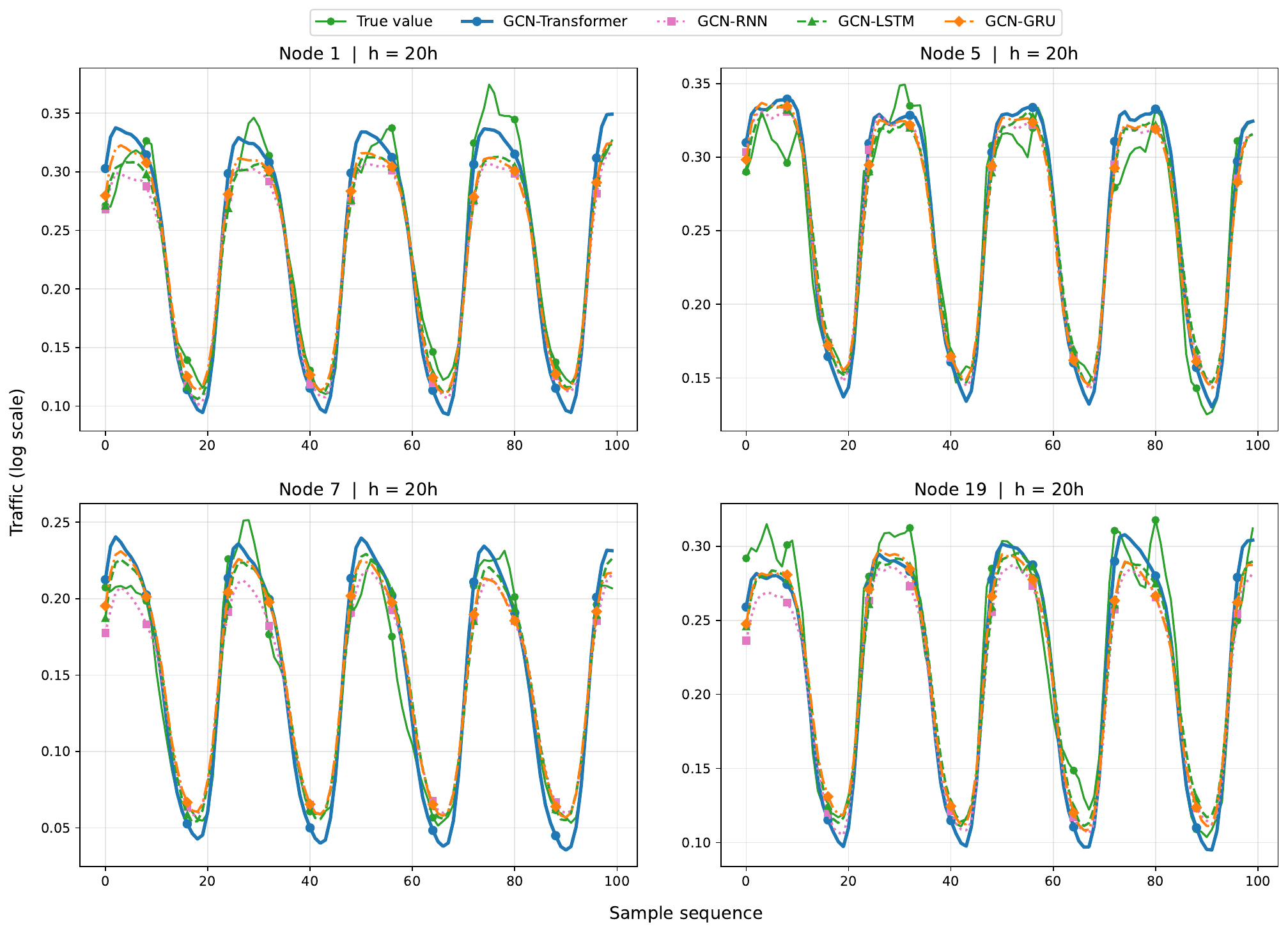}
 \caption{Predicted and actual traffic workload trajectories for four representative spatial nodes at prediction horizon $h{=}20$. Values are shown on the log scale over sampled test sequences; closer overlap with the true curve indicates better long-horizon tracking.}
 \label{fig:fig_sample_pred_4panel}
\end{figure*}

Fourth, session-level records are aggregated into hourly cluster-level traffic demand. Sessions spanning multiple hourly slots are split according to their overlap with each slot. The slot-level records are then aggregated by \texttt{cluster\_id} and \texttt{time\_slot} to compute traffic demand indicators such as the number of unique users, the number of connections, and total workload. Missing cluster-time combinations are completed to form a regular spatiotemporal panel. Additional temporal and traffic workload features, including hour of day, day of week, weekend indicator, cyclic time features, average workload per site, and average session duration, are generated for model input.

\subsubsection{Graph and Model Settings}

A fully connected spatial graph is constructed over the 25 cluster centroids. Each node corresponds to one edge service region, and each pair of distinct clusters is connected by a distance-weighted edge. Consistent with the graph construction in Section~\ref{sec_method}, edge weights are computed from centroid distances and normalized to form the adjacency matrix used by the graph convolutional module. The same fixed geographic graph is used for the proposed model and all baselines.

Table~\ref{tab:model-hyperparameters} summarizes the main model settings. All models use a hidden dimension of 64, a 168-hour historical input window, a 24-hour forecasting horizon, and a dropout rate of 0.2. The Transformer model uses four attention heads. These shared settings provide a controlled comparison among different temporal modeling modules.

\begin{table}[htbp]
\centering
\caption{Hyperparameter settings of the compared models.}
\label{tab:model-hyperparameters}
\resizebox{\columnwidth}{!}{
\begin{tabular}{lccccc}
\hline
\textbf{Model} & \textbf{Hidden} & \textbf{Heads} & \textbf{Window} & \textbf{Horizon} & \textbf{Dropout} \\
\hline
GCN-Transformer & 64 & 4 & 168 h & 24 h & 0.2 \\
GCN-RNN & 64 & -- & 168 h & 24 h & 0.2 \\
GCN-LSTM & 64 & -- & 168 h & 24 h & 0.2 \\
GCN-GRU & 64 & -- & 168 h & 24 h & 0.2 \\
\hline
\end{tabular}
}
\end{table}

\subsubsection{Model Implementation}

The proposed model follows the architecture described in Section~\ref{sec_method}. After preprocessing, supervised samples are generated with a sliding-window strategy. For each mini-batch, the input tensor has shape $B\times W\times N\times F$, where $B$ is the batch size, $W$ is the historical window length, $N$ is the number of spatial clusters, and $F$ is the number of node-level features. The prediction target has shape $B\times H\times N$, where $H$ is the prediction horizon.

The spatial encoder uses two graph convolution layers to propagate information over the normalized adjacency matrix. In addition to the fixed geographic graph, a learnable adaptive adjacency component is included to capture possible data-driven relationships not fully represented by distance alone. The temporal module uses a two-layer Transformer encoder with sinusoidal positional encoding. Temporal attention pooling summarizes the encoded sequence, and a regression head produces multi-horizon traffic demand forecasts for all clusters.

\subsubsection{Baseline Models}

The proposed GCN-Transformer is compared with three recurrent graph-based baselines that share the same spatial encoder but use different temporal modules. \textbf{GCN-RNN} uses a standard Elman recurrent network, \textbf{GCN-LSTM} uses a long short-term memory network, and \textbf{GCN-GRU} uses a gated recurrent unit network. All baselines use the same hidden dimension, input window, prediction horizon, dropout rate, optimizer, and data split as the proposed model. This design keeps the spatial representation fixed and isolates the effect of temporal modeling.

\subsection{Forecasting Results}

\subsubsection{Training and Testing Behavior}

\begin{figure}[t]
 \centering
 \includegraphics[width=3.5in]{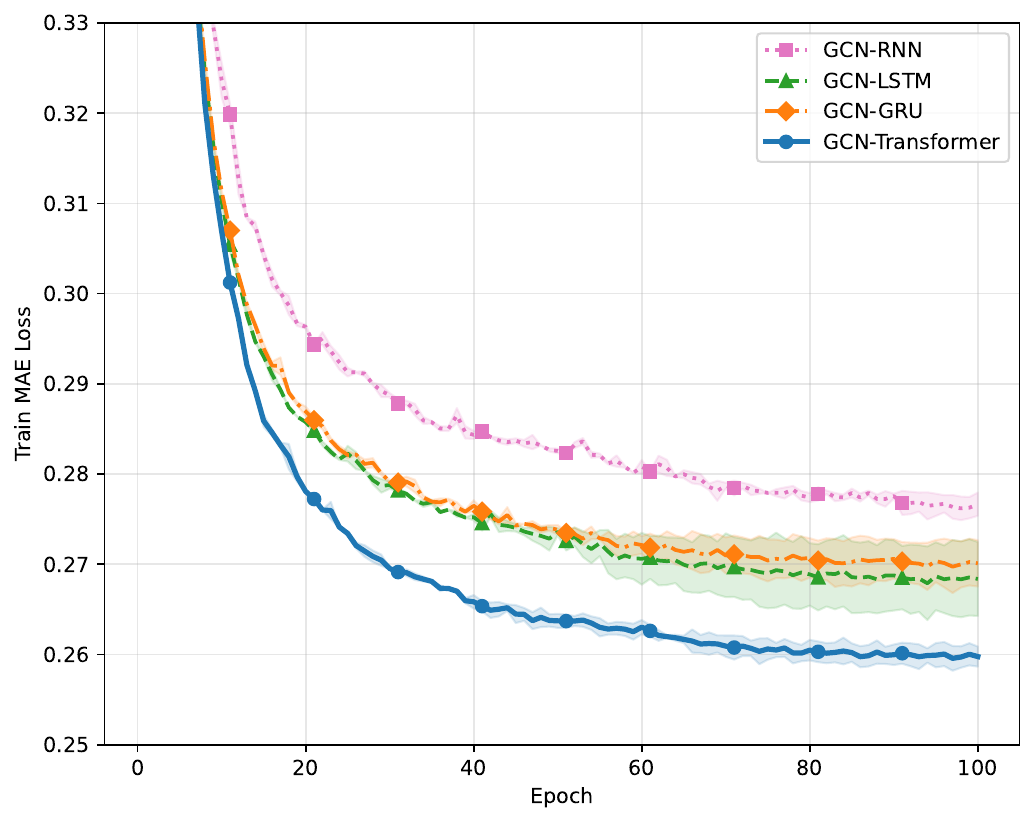}
 \caption{Training MAE loss of the compared models over 100 epochs. Lower values indicate smaller multi-horizon prediction errors on the training set.}
 \label{fig:train_loss}
\end{figure}

\begin{figure}[t]
 \centering
 \includegraphics[width=3.5in]{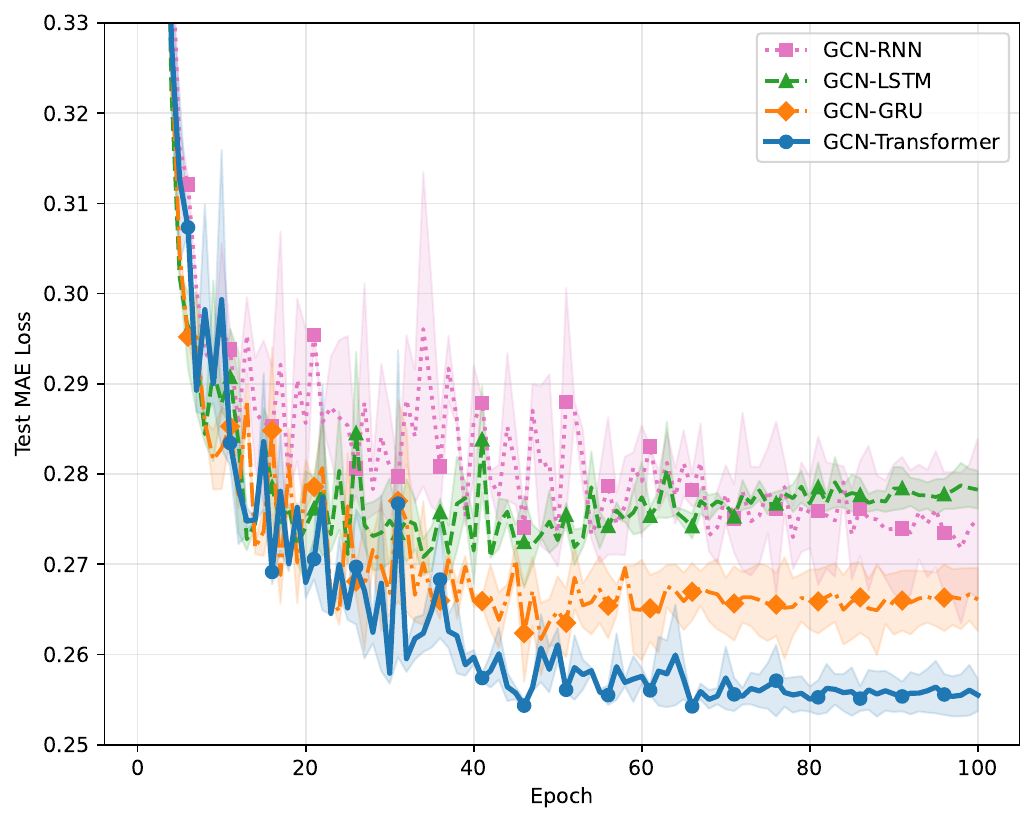}
 \caption{Testing MAE loss of the compared models over 100 epochs. Lower and more stable curves indicate better generalization under the selected test split.}
 \label{fig:test_loss}
\end{figure}

Figures~\ref{fig:train_loss} and~\ref{fig:test_loss} compare the training and testing MAE losses of the four models over 100 epochs. All models reduce the training loss rapidly during the early epochs and then converge more gradually. The GCN-Transformer reaches the lowest training loss and also maintains the lowest testing loss after convergence, indicating that its improvement is not limited to fitting the training samples. GCN-GRU provides the strongest recurrent baseline, whereas GCN-RNN and GCN-LSTM converge to higher and more fluctuating testing losses. These curves suggest stable optimization and provide a consistent basis for the horizon-wise accuracy comparison.

\subsubsection{Multi-Horizon Accuracy}

\begin{figure}
 \centering
 \includegraphics[width=3.5in]{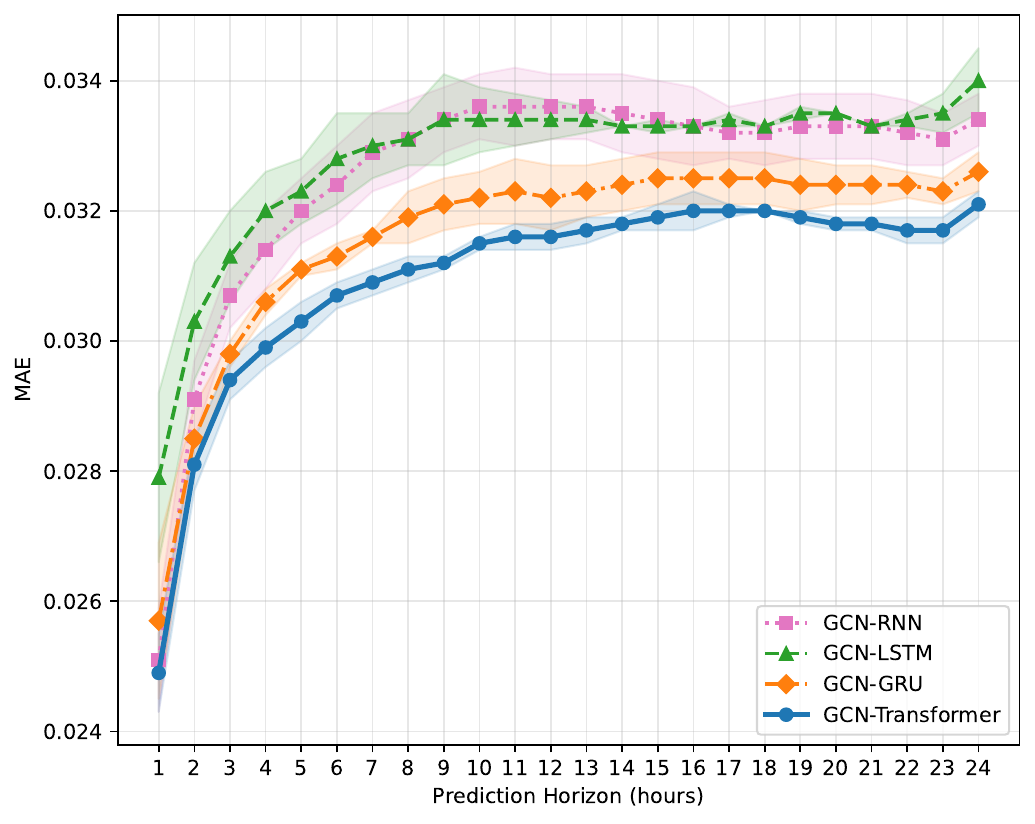}
 \caption{Horizon-wise MAE comparison of GCN-RNN, GCN-LSTM, GCN-GRU, and GCN-Transformer over the 1--24 h forecasting range. Lower MAE indicates higher forecasting accuracy.}
 \label{fig:comparison_mae}
\end{figure}

Figure~\ref{fig:comparison_mae} compares the mean absolute error (MAE) of all models from 1-hour to 24-hour prediction horizons. For all models, MAE increases quickly over the first several horizons and then changes more gradually, confirming that longer forecasts are more difficult. The GCN-Transformer achieves the lowest MAE at every plotted horizon. Its advantage over GCN-GRU is moderate but consistent, while its advantage over GCN-RNN and GCN-LSTM becomes more visible at medium and long horizons. This pattern suggests that self-attention helps the model retain temporally relevant information from the historical window, rather than depending only on recurrent state propagation.

\subsubsection{Spatial Residual Analysis}

\begin{figure}
 \centering
 \includegraphics[width=3.5in]{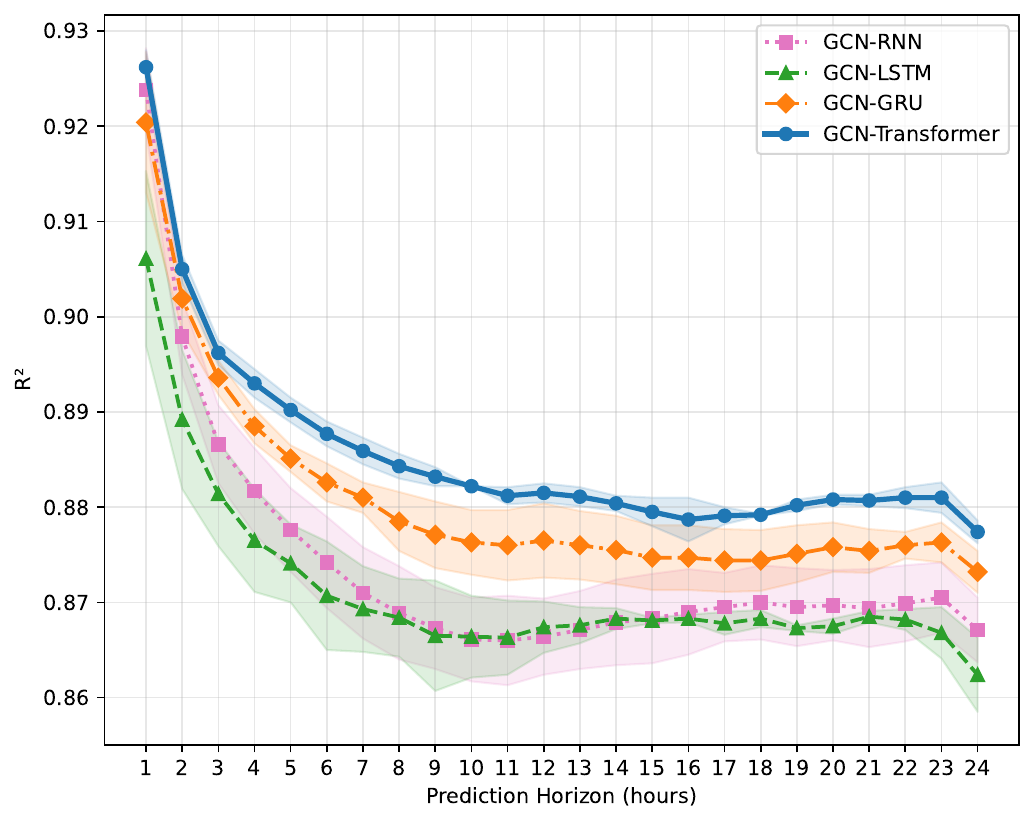}
 \caption{Horizon-wise $R^2$ comparison of GCN-RNN, GCN-LSTM, GCN-GRU, and GCN-Transformer over the 1--24 h forecasting range. Higher $R^2$ indicates greater explained variance.}
 \label{fig:comparison_r2}
\end{figure}

\begin{figure*}
 \centering
 \includegraphics[width=7.5in]{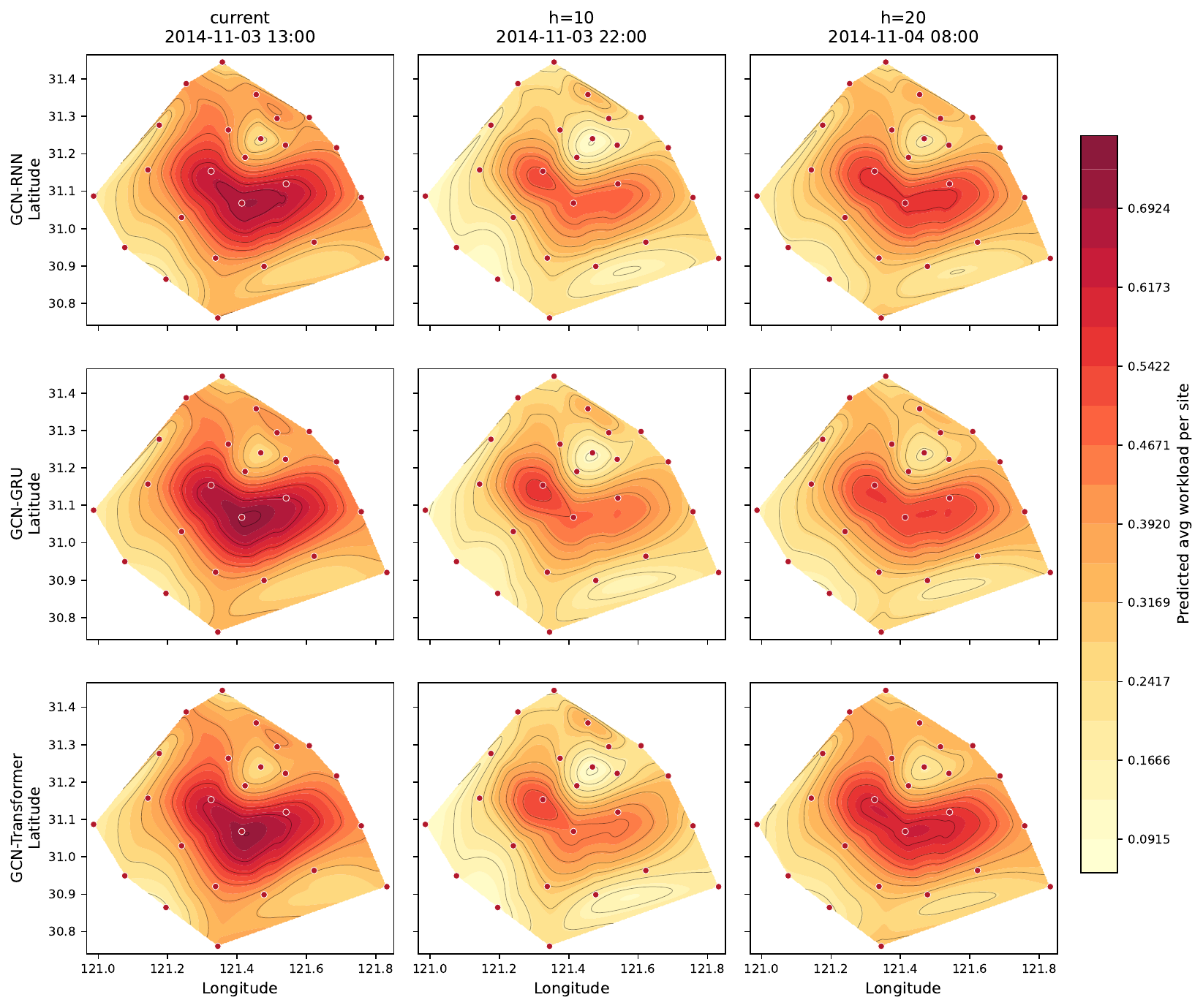}
 \caption{Spatial distribution of predicted average traffic workload per site for GCN-RNN, GCN-GRU, and GCN-Transformer. Rows correspond to models and columns correspond to the current timestamp, $h{=}10$, and $h{=}20$; warmer colors indicate higher predicted traffic workload and dots mark cluster centroids.}
 \label{fig:prediction_heatmap_current}
\end{figure*}

Figure~\ref{fig:residual_heatmap_current} shows the spatial distribution of prediction residuals, defined as prediction minus ground truth, for GCN-RNN, GCN-GRU, and GCN-Transformer at the current timestamp, $h{=}10$, and $h{=}20$. Red regions indicate over-prediction, whereas blue regions indicate under-prediction. At the current timestamp, all models show mixed residuals with localized positive and negative errors. At $h{=}10$, the recurrent baselines exhibit broader positive residuals in the eastern part of the service area and stronger negative residuals in the northwest, indicating more spatially coherent bias. The GCN-Transformer still has local errors, but its residuals are less concentrated. At $h{=}20$, the recurrent baselines show wider underestimation regions, while the GCN-Transformer produces a more balanced residual field. These maps suggest that the Transformer improves not only aggregate error, but also the spatial distribution of prediction errors, although persistent residual patches show that long-horizon spatial forecasting remains challenging.

Overall, the results indicate that combining graph-based spatial aggregation with Transformer-based temporal memory improves forecasting stability and long-horizon accuracy on the evaluated dataset. The improvement is most visible when the forecast horizon increases and when traffic demand transitions occur across time. These results support the use of attention-based temporal memory for traffic-aware proactive edge resource provisioning, while also indicating that further gains may be possible through richer contextual features or adaptive graph construction.

\subsection{Additional Evaluation Results}
\label{sec_additional_results}

In addition to the main MAE and residual analyses, this subsection checks whether the same model ranking remains visible under explained variance, node-level trajectories, and predicted spatial traffic workload fields.

\subsubsection{$R^2$ Performance Comparison}

Figure~\ref{fig:comparison_r2} reports the $R^2$ scores of all four models over the 24-hour prediction horizon. Higher $R^2$ indicates that a model explains more of the variance in the observed traffic workload. For all models, $R^2$ decreases from short to medium horizons and then stabilizes, which is consistent with the increasing difficulty of long-horizon forecasting. The GCN-Transformer maintains the highest $R^2$ across all horizons, while GCN-GRU provides the strongest recurrent baseline. GCN-RNN and GCN-LSTM obtain lower long-horizon scores, with GCN-LSTM showing the largest degradation near the final horizon. This result is consistent with the MAE comparison and confirms that the observed improvement is not specific to a single error metric.

\subsubsection{Per-Node Prediction Trajectories}

Figure~\ref{fig:fig_sample_pred_4panel} compares predicted and actual traffic workload trajectories for four representative spatial nodes at prediction horizon $h{=}20$. All models capture the dominant periodic pattern, indicating that the historical window contains useful temporal regularity. The differences appear mainly in the amplitude and timing of peaks and troughs. GCN-RNN tends to damp some high-traffic periods, while GCN-GRU and GCN-LSTM provide stronger recurrent predictions. The GCN-Transformer often follows the phase and amplitude of the true trajectory more closely, especially around repeated daily transitions, but some abrupt peaks remain smoothed. This indicates that the proposed model improves long-horizon tracking while leaving room for future improvements in sharp peak prediction.

\subsubsection{Spatial Prediction Analysis}

Figure~\ref{fig:prediction_heatmap_current} visualizes the predicted average traffic workload at the current timestamp, $h{=}10$, and $h{=}20$. All models identify the main high-traffic region near the central part of the service area, which shows that the shared graph encoder captures the dominant spatial structure. Differences appear in how each model preserves the intensity and gradients of this hotspot across horizons. The GCN-Transformer produces a more coherent high-traffic region and smoother surrounding gradients, especially at the longer horizon. The recurrent baselines capture the general location of traffic demand but show more attenuation or distortion in some regions. This spatial prediction view supports the residual analysis and indicates that temporal modeling affects the quality of spatial traffic workload fields, even when the graph encoder is shared.

\section{Conclusions}
\label{sec_con}

In this paper, we investigated traffic intelligence as a spatiotemporal learning problem for edge computing systems. To address the joint challenges of spatially coupled service regions and long-horizon traffic dynamics, we proposed a graph-based Transformer memory framework that separates spatial perception from temporal reasoning. The graph convolutional component captures correlations among edge service regions through geographic and adaptive adjacency relationships, while the Transformer-based memory module models temporal dependencies across historical traffic workload sequences.

Experimental results on a real cellular network dataset demonstrate that the proposed GCN-Transformer consistently outperforms recurrent graph-based baselines, including GCN-RNN, GCN-LSTM, and GCN-GRU. The advantage is especially clear at longer prediction horizons, where recurrent models are more likely to lose important historical information or underestimate traffic recovery. The residual and spatial prediction analyses further show that the Transformer memory module produces more coherent spatial traffic patterns and better captures transitions between low-activity and high-activity periods.

These findings indicate that attention-based temporal memory is a useful mechanism for traffic-aware proactive edge resource management. More accurate long-horizon traffic forecasting can help edge controllers allocate resources ahead of traffic changes, reduce overload risk, and improve the reliability of latency-sensitive services. Future work will extend this framework by incorporating additional contextual information, such as mobility traces, service types, and event-driven traffic changes, and by integrating the forecasting model with online resource allocation policies in dynamic edge environments.

\nocite{*}
\bibliographystyle{IEEEannot}
\bibliography{annot}

\begin{thebibliography}{10}
\providecommand{\url}[1]{#1}
\csname url@rmstyle\endcsname
\providecommand{\newblock}{\relax}
\providecommand{\bibinfo}[2]{#2}
\providecommand\BIBentrySTDinterwordspacing{\spaceskip=0pt\relax}
\providecommand\BIBentryALTinterwordstretchfactor{4}
\providecommand\BIBentryALTinterwordspacing{\spaceskip=\fontdimen2\font plus
\BIBentryALTinterwordstretchfactor\fontdimen3\font minus \fontdimen4\font\relax}
\providecommand\BIBforeignlanguage[2]{{%
\expandafter\ifx\csname l@#1\endcsname\relax
\typeout{** WARNING: IEEEtran.bst: No hyphenation pattern has been}%
\typeout{** loaded for the language `#1'. Using the pattern for}%
\typeout{** the default language instead.}%
\else
\language=\csname l@#1\endcsname
\fi
#2}}

\bibitem{Xu2021}
D.~Xu, T.~Li, Y.~Li, X.~Su, S.~Tarkoma, T.~Jiang, J.~Crowcroft, and P.~Hui, ``Edge intelligence: Empowering intelligence to the edge of network,'' \emph{Proceedings of the IEEE}, vol. 109, no.~11, pp. 1778--1837, 2021.


\bibitem{10944303}
J.~Peng, H.~Xing, X.~Chen, Y.~Li, Y.~Cui, D.~Zheng, L.~Ale, and L.~Feng, ``Security enhanced computation offloading for collaborative inference at semantic-communication-empowered edge,'' \emph{IEEE Transactions on Mobile Computing}, vol.~24, no.~9, pp. 8071--8088, 2025.


\bibitem{Ale2024}
\BIBentryALTinterwordspacing
L.~Ale, N.~Zhang, S.~A. King, and D.~Chen, ``Empowering generative {AI} through mobile edge computing,'' \emph{Nature Reviews Electrical Engineering}, vol.~1, no.~7, pp. 478--486, 2024. [Online]. Available: \url{http://dx.doi.org/10.1038/s44287-024-00053-6}
\BIBentrySTDinterwordspacing


\bibitem{Patel2014}
M.~Patel, Y.~Hu, P.~H{\'e}d{\'e}, J.~Joubert, C.~Thornton, B.~Naughton, \emph{et~al.}, ``Mobile-edge computing - introductory technical white paper,'' Mobile-edge Computing (MEC) industry initiative,'' White Paper, Sep. 2014.


\bibitem{Deng2021}
S.~Deng, Z.~Xiang, J.~Taheri, M.~A. Khoshkholghi, J.~Yin, A.~Y. Zomaya, and S.~Dustdar, ``Optimal application deployment in resource constrained distributed edges,'' \emph{IEEE Transactions on Mobile Computing}, vol.~20, pp. 1907--1923, May. 2021.


\bibitem{Sun2022MobileData}
F.~Sun, P.~Wang, J.~Zhao, N.~Xu, J.~Zeng, J.~Tao, K.~Song, C.~Deng, J.~C. Lui, and X.~Guan, ``Mobile data traffic prediction by exploiting time-evolving user mobility patterns,'' \emph{IEEE Transactions on Mobile Computing}, vol.~21, no.~12, pp. 4456--4470, 2022.


\bibitem{Wu2024}
T.~Wu, X.~Fan, H.~Wei, Y.~Qu, C.~Xiang, P.~Yang, and F.~Wu, ``Predictive service provisioning with online learning in wireless edge networks,'' \emph{IEEE Transactions on Mobile Computing}, vol.~23, no.~5, pp. 4076--4091, 2024.


\bibitem{Duc2019}
\BIBentryALTinterwordspacing
T.~L. Duc, R.~G. Leiva, P.~Casari, and P.-O. \"{O}stberg, ``Machine learning methods for reliable resource provisioning in edge-cloud computing: A survey,'' \emph{ACM Comput. Surv.}, vol.~52, no.~5, Sept. 2019. [Online]. Available: \url{https://doi.org/10.1145/3341145}
\BIBentrySTDinterwordspacing


\bibitem{Zhang2023}
\BIBentryALTinterwordspacing
X.~Zhang and S.~Debroy, ``Resource management in mobile edge computing: A comprehensive survey,'' \emph{ACM Comput. Surv.}, vol.~55, no. 13s, July 2023. [Online]. Available: \url{https://doi.org/10.1145/3589639}
\BIBentrySTDinterwordspacing


\bibitem{Ale2021}
L.~Ale, N.~Zhang, S.~A. King, and J.~Guardiola, ``Spatio-temporal bayesian learning for mobile edge computing resource planning in smart cities,'' \emph{ACM Transactions on Internet Technology}, vol.~21, Aug. 2021.


\bibitem{zhangTIDES2025}
C.~Zhang, H.~Zhang, J.~Qiao, Z.~Li, and M.-S. Alouini, ``Tides: Traffic intelligence with deepseek-enhanced spatial–temporal prediction,'' \emph{IEEE Journal on Selected Areas in Communications}, vol.~44, pp. 2544--2558, 2026.


\bibitem{zhangGradient2025}
C.~Zhang, H.~Zhang, S.~Dang, B.~Shihada, and M.-S. Alouini, ``Gradient compression and correlation driven federated learning for wireless traffic prediction,'' \emph{IEEE Transactions on Cognitive Communications and Networking}, vol.~11, no.~4, pp. 2246--2258, 2025.


\bibitem{Ale8660445}
L.~Ale, N.~Zhang, H.~Wu, D.~Chen, and T.~Han, ``Online proactive caching in mobile edge computing using bidirectional deep recurrent neural network,'' \emph{IEEE Internet of Things Journal}, vol.~6, no.~3, pp. 5520--5530, 2019.


\bibitem{Wang2017Spatiotemporal}
J.~Wang, J.~Tang, Z.~Xu, Y.~Wang, G.~Xue, X.~Zhang, and D.~Yang, ``Spatiotemporal modeling and prediction in cellular networks: A big data enabled deep learning approach,'' in \emph{IEEE INFOCOM 2017 - IEEE Conference on Computer Communications}, 2017, pp. 1--9.


\bibitem{Yao2023}
Y.~Yao, B.~Gu, Z.~Su, and M.~Guizani, ``{MVSTGN}: A multi-view spatial-temporal graph network for cellular traffic prediction,'' \emph{IEEE Transactions on Mobile Computing}, vol.~22, no.~5, pp. 2837--2849, May 2023.


\bibitem{Xu2017}
\BIBentryALTinterwordspacing
F.~Xu, Y.~Li, H.~Wang, P.~Zhang, and D.~Jin, ``Understanding mobile traffic patterns of large scale cellular towers in urban environment,'' \emph{IEEE/ACM Trans. Netw.}, vol.~25, no.~2, p. 1147–1161, Apr. 2017. [Online]. Available: \url{https://doi.org/10.1109/TNET.2016.2623950}
\BIBentrySTDinterwordspacing


\bibitem{Zhang2018}
\BIBentryALTinterwordspacing
C.~Zhang and P.~Patras, ``Long-term mobile traffic forecasting using deep spatio-temporal neural networks,'' in \emph{Proceedings of the Eighteenth ACM International Symposium on Mobile Ad Hoc Networking and Computing}, ser. Mobihoc '18.\hskip 1em plus 0.5em minus 0.4em\relax New York, NY, USA: Association for Computing Machinery, 2018, p. 231–240. [Online]. Available: \url{https://doi.org/10.1145/3209582.3209606}
\BIBentrySTDinterwordspacing


\bibitem{Elman1990}
J.~L. Elman, ``Finding structure in time,'' \emph{Cognitive Science}, vol.~14, pp. 179--211, Mar. 1990.


\bibitem{Hochreiter1997}
S.~Hochreiter and J.~Schmidhuber, ``Long short-term memory,'' \emph{Neural Computation}, vol.~9, no.~8, pp. 1735--1780, 1997.


\bibitem{Chung2014}
\BIBentryALTinterwordspacing
J.~Chung, C.~Gulcehre, K.~Cho, and Y.~Bengio, ``Empirical evaluation of gated recurrent neural networks on sequence modeling,'' Sep. 2014. [Online]. Available: \url{http://arxiv.org/abs/1412.3555}
\BIBentrySTDinterwordspacing


\bibitem{Wu2019}
\BIBentryALTinterwordspacing
Z.~Wu, S.~Pan, G.~Long, J.~Jiang, and C.~Zhang, ``Graph wavenet for deep spatial-temporal graph modeling,'' in \emph{Proceedings of the Twenty-Eighth International Joint Conference on Artificial Intelligence, {IJCAI-19}}.\hskip 1em plus 0.5em minus 0.4em\relax International Joint Conferences on Artificial Intelligence Organization, 7 2019, pp. 1907--1913. [Online]. Available: \url{https://doi.org/10.24963/ijcai.2019/264}
\BIBentrySTDinterwordspacing


\bibitem{Vaswani2017}
A.~Vaswani, N.~Shazeer, N.~Parmar, J.~Uszkoreit, L.~Jones, A.~N. Gomez, {\L}.~Kaiser, and I.~Polosukhin, ``Attention is all you need,'' in \emph{Advances in Neural Information Processing Systems}, vol.~30, 2017, pp. 5998--6008.


\bibitem{Zhou2021Informer}
\BIBentryALTinterwordspacing
H.~Zhou, S.~Zhang, J.~Peng, S.~Zhang, J.~Li, H.~Xiong, and W.~Zhang, ``Informer: Beyond efficient transformer for long sequence time-series forecasting,'' \emph{Proceedings of the AAAI Conference on Artificial Intelligence}, vol.~35, no.~12, p. 11106–11115, May 2021. [Online]. Available: \url{https://ojs.aaai.org/index.php/AAAI/article/view/17325}
\BIBentrySTDinterwordspacing


\bibitem{Gu2023Spatial-TemporalTransformer}
B.~Gu, J.~Zhan, S.~Gong, W.~Liu, Z.~Su, and M.~Guizani, ``A spatial-temporal transformer network for city-level cellular traffic analysis and prediction,'' \emph{IEEE Transactions on Wireless Communications}, vol.~22, no.~12, pp. 9412--9423, 2023.


\bibitem{Hu2023Citywide}
Y.~Hu, Y.~Zhou, J.~Song, L.~Xu, and X.~Zhou, ``Citywide mobile traffic forecasting using spatial-temporal downsampling transformer neural networks,'' \emph{IEEE Transactions on Network and Service Management}, vol.~20, no.~1, pp. 152--165, 2023.


\bibitem{Gong2025STTF}
J.~Gong, Y.~Liu, T.~Li, J.~Ding, Z.~Wang, and D.~Jin, ``Sttf: A spatiotemporal transformer framework for multi-task mobile network prediction,'' \emph{IEEE Transactions on Mobile Computing}, vol.~24, no.~5, pp. 4072--4085, 2025.


\bibitem{Scarselli2009}
F.~Scarselli, M.~Gori, A.~C. Tsoi, M.~Hagenbuchner, and G.~Monfardini, ``The graph neural network model,'' \emph{IEEE Transactions on Neural Networks}, vol.~20, pp. 61--80, Jan. 2009.


\bibitem{He2022}
K.~He, X.~Chen, Q.~Wu, S.~Yu, and Z.~Zhou, ``Graph attention spatial-temporal network with collaborative global-local learning for citywide mobile traffic prediction,'' \emph{IEEE Transactions on Mobile Computing}, vol.~21, no.~4, pp. 1244--1256, 2022.


\bibitem{Wang2017}
X.~Wang, Z.~Zhou, F.~Xiao, K.~Xing, Z.~Yang, Y.~Liu, and C.~Peng, ``Spatio-temporal analysis and prediction of cellular traffic in metropolis,'' \emph{IEEE Transactions on Mobile Computing}, vol.~18, no.~9, pp. 2190--2202, 2019.


\end{thebibliography}

\end{document}